# Event-Driven Models

Dimiter Dobrev
Institute of Mathematics and Informatics
Bulgarian Academy of Sciences
*d@dobrev.com*

In Reinforcement Learning we look for meaning in the flow of input/output information. If we do not find meaning, the information flow is not more than noise to us. Before we are able to find meaning, we should first learn how to discover and identify objects. What is an object? In this article we will demonstrate that an object is an event-driven model. These models are a generalization of action-driven models. In Markov Decision Process we have an action-driven model which changes its state at each step. The advantage of event-driven models is their greater sustainability as they change their states only upon the occurrence of particular events. These events may occur very rarely, therefore the state of the event-driven model is much more predictable.

## Introduction

What is an object? An object can be your favourite song, a single word or a semantic category (such verb or noun). An object can also be a person, an inanimate item or an animal, it can be your house or the whole neighbourhood you live in.

We are all able to identify objects. You will recognise your favourate song even if you hear only a short fragment of it. You may hear it performed by another singer or even from an audio stream of poor quality, but you will still know it is your favourate song.

In your mind the objects are structured according to some hierarchy, such as animal–dog–poodle –your poodle. Some objects possess the property *uniqueness*. E.g. your dog is unique and if you groom it will then be a groomed dog. If you groomed a random dog this will not imply that all dogs have been groomed.

*Uniqueness* will not be an invariable feature. You may believe that a certain person is unique, but all of a sudden you find out that he or she has a twin. For years on end you thought you were talking to the same person and now it turns out that you have talked to two different persons. Another example: There is a chair in your house and in your eyes this chair is unique. You certainly know that thousands of this type of chair have been produced, but your chair is the only one of these chairs in your house. If you repaint the chair you will expect it to be repainted the next time you see it. It may turn out that there are two such chairs in your house and you have repainted one of them.

How do we recognise (identify) objects? You may hear the beginning, the middle or the end of your favourite song. You may see a person you know in profile or in full face. You can explore a neighbourhood by starting from many different junctions. Thus, in order to recognise an object, we need not see it each time in the way we have seen it already.

An oriented graph is also an object and we recognise it by following a certain path in that graph. Of course the path must be sufficiently specific to that oriented graph, otherwise we cannot be sure it is exactly the graph we mean. If we are not sure, we can make an experiment by turning in a direction which would provide more information. Let us say you are at some street crossing, but do not know if the crossing is in your neighbourhood. So you take to the main street in order to find your bearings. Or you see someone you probably know, but are not sure. Then you can take a look at that person from all sides or shout out his name to see if he will look your way.

But we are not always in control of the situation or able to make an experiment. Sometimes we cannot hear well a spoken word. We can ask the speaker to repeat it, but in most cases it is a done deal and the party is over.

Objects in this article will be presented as oriented graphs (event-driven models). The vast majority of objects are not in front of eyes all the time, i.e. they appear and vanish from sight. Are there objects which we observe all the time? Are there oriented graphs in which we stay all the time and never leave? Yes, there are such objects (models). One example are the days of the week. They give us a model with seven states (Sunday to Saturday) and at any point of time we are in one of these states. Another example is our neighbourhood. If we spend all our life in that neighbourhood without ever leaving it, the neighbourhood will be permanently within our sight.

Most event-driven models will have a special state which we will term *outside*. An object will be off our sight when we are in an *outside* state.

The key question with event-driven models is "Where am I now?", i.e. in which state am I? If the event-driven model is an object, the key question is "Do I see the object now?". In other words, am I in the *outside* state or in some of the other states?

We will begin this article by presenting an intuitive idea of event-driven models. Then we will make a comparison between event-driven models and action-driven models. Next, we will provide an informal description of the problem and will explain why we will not try to find the initial state. The term *history* will be defined and the objective of the agent will be announced. We will demonstrate that we can have a variety of criteria and will challenge Sutton's assertion in [2] that there is only a single criterion. We will prove that the term *discount factor* must not be part of the definition of Reinforcement Learning (RL), because the *discount factor* is part of the strategy rather than of the meaning of life. We will continue with a discourse on whether the model of the world should be deterministic or nondeterministic. As regards incorrect moves, we will demonstrate that we can do without them, but should preferably assume that incorrect moves do exist. A definition of RL will be provided. Then we will define the *perfect model*, which is an action-driven model. We will then complicate the model by adding randomness. We will discard the assumption that random values have any distribution, and will demonstrate that such an assumption is wrong. We will point out that there must be a *trace*, i.e. some particular occurrence which distinguishes one state from another. Next, we will introduce the concept of *exhaustive models* and will demonstrate that they are as unattainable as the perfect ones. We will explain what is an event and will proceed with yet another complication of the model which will transform it from action-driven to event-driven. In fact we will replace actions with random events. We will add variables to the model. Finally we will create the Cartesian product of all adequate models discovered. That Cartesian product will be the model which we are looking for and which provides the best possible description of the world.

## The intuitive idea

Before we entangle the reader in a swarm of words, we will try to offer an intuitive idea of what the event-driven model is. This is an oriented graph similar to the ones shown in Figures 1 and 2.

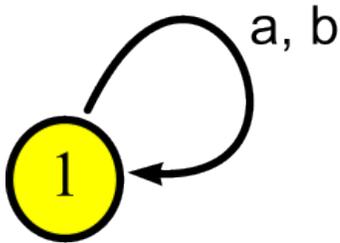

Figure 1

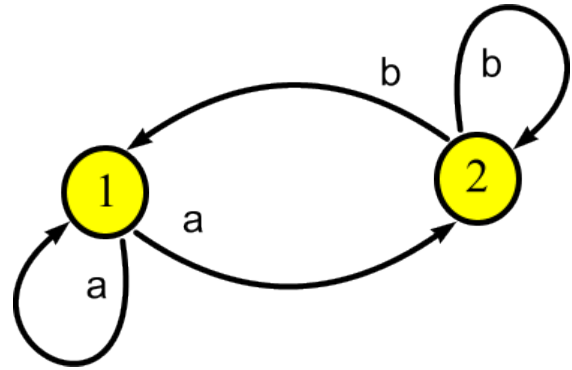

Figure 2

Here *a* and *b* are events. If *a* and *b* were actions, then Figures 1 and 2 would depict action-driven models. An action is of course an event, too. So action-driven models are a special case of event-driven models.

The model in Figure 1 is fairly simple and if we gathered statistics from it all we would get to know is how many times event *a* and respectively event *b* has occurred. In turn, that would indicate which event is more likely to occur.

The model in Figure 2 is more interesting. Here, if we are in state 1, event *b* cannot happen. The same applies to state 2 and event *a*. This model therefore *predicts* the next event (*a* or *b*). If we know the state we are in, we will also know which *the next event will be*.

If we flip the arrows (Figure 3), we will get a model which *remembers* the last event (*a* or *b*). If we know the state we are in, we will also know which event *was the last to occur*.

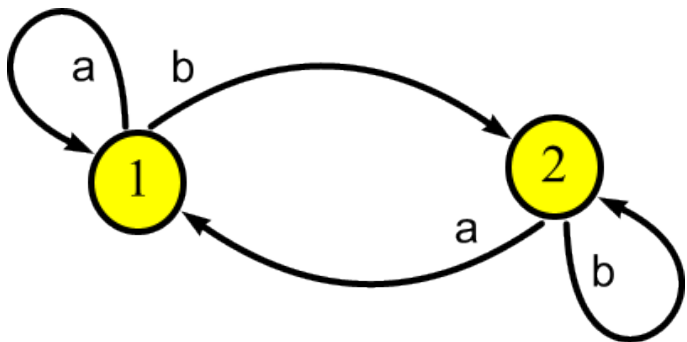

Figure 3

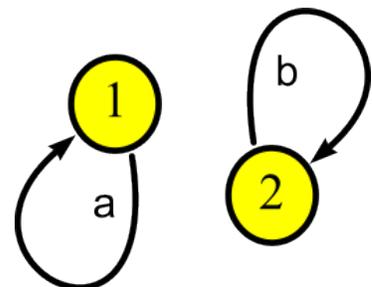

Figure 4

The commonality between Figure 2 and Figure 3 is that in either case there must be some difference between states 1 and 2. Something specific must occur in these states. We will name this specific occurrence *trace*. In the absence of a trace, i.e. of something specific, these two modes are useless.

A trace is indispensable. Each and every oriented graph is a model of the world (events correspond to arrows). If there were no trace (i.e. in the absence of a specific occurrence), the model would be inadequate and useless.

In Figure 2, the difference between states 1 and 2 is that state 1 has only outbound *a* arrows (and state 2 has only outbound *b* arrows). If this is all the difference they have (i.e. if there is no trace) we will never know whether we are in state 1 or in state 2, because when viewed from the perspective of the past the two states are indistinguishable.

In Figure 3 we know exactly the state we are in, but this would be useless if these states are indistinguishable from the viewpoint of the future (i.e. if there is no trace). In other words, it would be absolutely meaningless to remember the last event (*a* or *b*) if it does not bear any consequence for the future.

Let us illustrate this with a criminal scenario (Figure 4). We are looking for the perpetrator of a murder. Who is the perpetrator, the postman or the tramp? If the postman is the murderer, we are in state 1. If the tramp is the murderer, we are in state 2. Event *a* is "We proved that the postman perpetrated the murder". Accordingly, event *b* is "We proved that the tramp perpetrated the murder". The question is, which state are we in. In other words, who is the murderer? We will find out only when we come across hard evidence (the event *a* or *b*). Unfortunately, we may never come across hard evidence. The point is how to predict who the murderer is before we get hold of hard evidence. We may hypothesize that if the murder was perpetrated by the postman, we are more likely to come across clues that confirm he is the perpetrator than to clues that acquit the postman. In other words, the clues do not expose the perpetrator, but set the crosshairs on the more likely perpetrator. The clues will be provided by the *trace* in this model.

The model in Figure 4 is not very much to our liking, because we need recurrence. We want to traverse each arrow multiple times. Figure 4 however is an either-or scenario because the occurrence of *a* forecloses the occurrence of *b*.

Recurrence is very important. Heraclitus said that no man steps in the same river twice, for it is a different river each time. Likewise, no one can enter the same room twice (or meet the same person twice). Each time the room is different. Somebody has painted the walls, somebody else has rearranged the furniture. Somebody may come in, somebody else may go out. If all things are unchanged, a butterfly will come by and spoil all recurrence.

Nonetheless, we wish to step in the same river or room many times and meet repeatedly one and the same person. They may not be exactly the same, but we are willing to disregard minor or even major differences for the sake of making the world more simple and comprehensible.

This is exactly the underlying idea of event-driven models. These are models with a few states, which we visit multiple times and each visit lasts long (consists of many steps). The *modus operandi* of action-driven models and in particular of perfect models is exactly the opposite. A perfect model always exists. The easiest way to construct such a model is to arrange the states in a row and traverse all these states one time only. The state in an action-driven model describes everything, which essentially prevents one state from occurring twice (same as we cannot step in the same river twice).

## Event-Driven vs Action-Driven

What is the difference between event-driven and action-driven models? In Markov Decision Process (MDP) the model is an oriented graph which changes its state after each action (i.e. at each step). This model resembles a machine which ticks too fast. It is fairly difficult to answer the question "Where am I now?" with such a model. Therefore, it is a huge challenge to tell the current state. Let us imagine that human being is a step device which makes 24 steps per second. (Cinematographic movies run at a rate of 24 frames per second, but we cannot see the intermittency and perceive the imagery as seamless. So 24 steps per second is a good proposition.) If your state was changing 24 times per second you would hardly be able to tell your current state. Now let us have a model which does not change at every action, but at the occurrence of a few more interesting events, e.g. sunrise and sunset. These events occur once in 24 hours. At sunrise we transition from night to day and vice versa at sunset. When I ask you if it is day or night now you will most probably be able to give the right answer. In other words, the current state of the event-driven model is far more predictable (because it is a much more stable and changes less often).

Another difference is that in action-driven models we can try describe the behaviour of the world without considering the behaviour of the agent, but the event-driven model requires us to describe the world and the agent living in it as a composite system.

The MDP for example calculates the probability of a certain transition for a certain state and action. It does not however calculate how probable is it for the agent to perform a certain action because it tries to describe only the world without the agent. In event-driven models we have to describe the world and the agent as a composite system since when an event occurs we cannot tell if it occurred only because the agent wanted it or because that's the way the world goes. In MDP we do not predict the past, because the choice of the arrow which takes us to a certain state depends not only on the world, but on the agent as well. For example (Figure 5), if we are in state 2 and the last action is a red arrow, then we may have come to state 2 from states 1 and 5. If we knew that in state 1 the agent will never choose a red arrow, then the previous state must have been 5.

In event-driven models we make no distinction between past and future. We collect statistics for both and try predict both.

Describing the world and the agent as a composite system is much easier than describing the world without the agent. By way of example, you will never dye your hair in green colour, which is a simple description of you and of the world you live in. The reason is irrelevant – it may be that you are reasonable enough not to do so or that green dye is not available. If someone asked "Will your hair be green tomorrow?" you may answer confidently that this cannot happen. How you know it – from the analysis of your own behaviour or because of some limitations from the world – does not matter. What matters is that your hair cannot be green tomorrow.

## Informal description

Initially we will provide an informal description of the problem. We have a world (environment) and an agent who lives in that world. The world is an oriented graph similar to the one depicted in Figure 5. The agent moves from one state to another by following the arrows. The states and the arrows are tagged with some labels. However, in Figure 5 we have used colour codes instead of labels. As the agent moves, he enters (observes) information (the label of his current state) and performs actions (chooses the label of the arrow of his next move). Note that the agent choses only the label (colour) of the arrow, but not the particular arrow. E.g. if from state 6 the agent choses a blue arrow, he cannot know whether the blue arrow will take him to state 5 or to state 7. Some choices may not be available, for example there are not outbound red arrows from state 2. So in state 2 the agent cannot choose to continue his path on a red arrow and must choose a blue one.

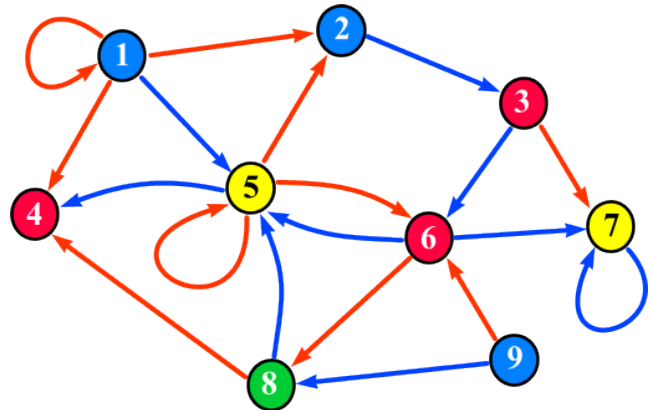

**Figure 5**

We will name the agent's path in the oriented graph *the life* of the agent. Any state can be the begging or the end of that life. However, there are states which – if they are part of the path – can only be the beginning of that path. We will term these states *absolute beginning*. These states do not have any inbound arrows, so there cannot be any past before them. An example in Figure 5 is state 9. State 1 is not an absolute beginning because of the red loop. Accordingly, we will term the states without outbound arrows *sudden death*. If such a state is part of the path, it must always be the last state. There is no future after such a state, cf. state 4 in Figure 5. State 7 is not sudden death because of the blue loop.

By naming the states without outbound arrow *sudden death* we mean that the agent's life may end in two ways – either we shut him down or he reaches a *sudden death* state and stops. We will say that shutting down the agent is natural death and when the agent stops on his own it is sudden death. (In everyday speech, when someone shuts us down we do not say it is natural death, but a murder. But in this article we assume that we will shut down the agent when much time has elapsed. So we will associate shutting down with dying of old age, therefore natural death.)

Note that life need not necessarily start from some absolute beginning. The past may be infinite in the same way as the future may be. Therefore, life can be an endless path which has no beginning and no end.

In MDP and RL we examine two scenarios. In one scenario the agent sees everything (Full Observability) and in the other scenario he sees only partially (Partial Observability). In Figure 5 we would have Full Observability if all the states had different colours or if the agent observed not the colour, but the number of states (i.e. if the agent sees which is the state he is in now).

Most articles suggest that in MDP we have Full Observability. This is the special case. They use the term Partially Observable Markov Decision Process (POMDP) to describe the generalization of Partial Observability. This article puts things the way round. We will use the term MDP to denote generalization, while for Full Observability will use the term Fully Observable Markov Decision Process (FO MDP).

Similarly, here we will assume that in Reinforcement Learning we have Partial Observability, and will flag explicitly the cases when we examine Full Observability.

## The initial state

In Reinforcement Learning (RL) we make references to an initial state and to a current state. In order to describe one path (life), we need a state to start from. Most authors tend to go for the initial state guided by the assumption that it is easier to predict forward in the direction of the future than backward in the direction of the past. Therefore, going in the direction of the arrows is easier than going against the arrows. In Markov Decision Process (MDP) for example we have probabilities which indicate what we expect the state to be if we move ahead with certain action, but there are no probabilities for the opposite direction. In this article we will not refer to initial states but only to current states. This comes from the assumption that the past and the future are on a level playing field and are equally difficult to predict.

There are many reasons why we will not be interested in the initial state. There may not be any initial state at all, because the past can be infinite. Which is the initial state of a human being? Is it the moment of birth or the moment of conception? In our model of the world there are moments which have occurred before our birth. Therefore, birth in our model is not an absolute beginning. There is one model of the world with an absolute beginning, and this is the Big Bang theory. Is that a sound theory? Rather than asking whether a theory is true or false, we need to know if the theory is fit for our purpose (does it describe the world). There is a raft of theories which describe the word. We will deem that a theory is true insofar as it is impossible to design an experiment which disproves that theory.

The questions we will ask in RL therefore are "What is the world?" and "Where are we now?". More often than not, we ask "Where am I now?" rather than "Where was I at the time I was born?". Consequently, we will look for the current state and not for the initial state.

## What is History?

What is life? Life is the whole history, i.e. a history which cannot continue either because it is infinite or because there is no next step, perhaps because the history has come to sudden death or because we have shut down the agent (natural death).

What is history? **Definition:** A *truncated history* will be the series of actions and observations from the initial moment to the current moment, including the last correct action.

$$a_1, v_1, a_2, v_2, \ldots, a_t, v_t, a_{t+1}$$

**Definition:** We will have *history* when we add all the incorrect (bad) moves which we have tried and which we know to be incorrect.

$$bad_0, a_1, v_1, bad_1, a_2, v_2, \ldots, bad_{t-1}, a_t, v_t, bad_t, a_{t+1}$$

In the above sequence *a* is action and *bad* is a set of actions. Note that each next step begins with the next correct move. The *bad* set can be regarded as part of the observations made at the previous step.

**Definition:** A *full history* is the result which we obtain when we replace the *subset* of incorrect (bad) moves which we tried with the *full set* of incorrect moves.

$$full_0, a_1, v_1, full_1, a_2, v_2, \ldots, full_{t-1}, a_t, v_t, full_t, a_{t+1}$$

What is the difference between *history* and *full history*? The first is what we have actually seen along the path and the latter is what we would have seen if were observing more attentively. So $bad \subseteq full$. That is, the moves which we have tried and know to be incorrect are part of all incorrect moves. We may keep trying more moves until $bad = full$, however, in doing so we may inadvertently play a correct move (different from $a_i$) which will make the history completely different.

**Definition:** A *local history* of length *k* will be the last *k* steps of a history. That is, the end of some history.

**Definition:** An *approximate history* will be some form of incomplete description of a history. Presumably, the history is too long and we will hardly be able to remember the whole of it. Thus, we will remember only the end of it, or a few more important events (the time of their occurrence), or some statistics about that history. Usually, we do not need the full history and an approximate history is sufficient to build a model of the world and plan our future actions.

Note that what matters for the agent are not the states he has traversed, but the observations he has made. While the life corresponds to a path in the oriented graph (the model of the world), two different paths may correspond to one life.

## The agent's objective

What is the agent's objective? His objective is to achieve a better life.

But which life is better? We need a relation which compares lives. This relation must be a quasiorder (preorder). That is, it must be both reflexive and transitive. If we add antisymmetry, we will get partial order. We do not, however, need antisymmetry, because it will not be a problem where two lives are equally successful.

We would like this relation to possess some monotonicity. Let us divide life in two halves and let us have two lives. If the first and the second halves of Life 1 are better respectively than the first and the second halves of Life 2, we would expect Life 1 to be better than Life 2 overall. The opposite would be bizarre.

The relation in question can be obtained by adding up rewards and regrets. But in this article instead of rewards and regrets we say *scores*. Rewards are positive scores, while regrets are negative scores. For some scores, though, we do not know whether they are positive or negative. Relativity is pervasive. For example, if you earn a *D* grade at school, is that positive or negative? For some students or parents it might be positive, but for others it may be disappointing.

Thus, summing up all the scores we will give a number which represents the overall score of the whole life. In [3] the number is equal to the sum of the scores, but life there has a fixed length and the sum equals the arithmetic mean. In [4] the number is calculated as an arithmetic mean of the scores. Later articles use a *discount factor*, which is a mistake and later we will discuss why this is a mistake.

Most articles imply that the agent gets a score at each step, e.g. [3, 4] and many others. This proposition can be seen as an attempt to simplify the discourse, but in principle the idea of having a score at each step is illogical and unnatural. At schools for example they do not examine each kid every day.

Another possible approach is to assume that if we do not have any score, then the score is zero, however all these zeros will alter significantly the arithmetic mean. That would equally alter our strategy. If for example we play chess and a draw is zero, then ending in a draw or continuing the game would be all the same. The length of the game will also be very important. Thus, if our mean score is negative, we will try to extend the game as much as possible. If the mean score is positive we will try to cut the game as short as possible. While it seems quite reasonable to try end the game as soon as possible, we would be prone to risk the victory for the sole sake of ending the game faster. At chess tournaments they calculate the arithmetic mean of all wins, losses and draws without counting the average number of moves per game.

Hence, while most articles imply that the score is a real number, we will suggest that the score can be either a real number or the constant *Undef*. Therefore, we will suppose that at each step we may or may not have a score.

## Different criteria

Most articles imply that we have only one criterion against which life can be measured. In economics this criterion is money and the better strategy is the one which reaps more profit. We will assume that there are two criteria: of course we want the highest profit, but without ending up in jail.

We will look at objectives which are defined by two or more criteria. Let us have two persons, one has more wealth, and the other has more children. Who has been more successful? If the two criteria are equally important, Life 1 must outperform Life 2 on both criteria in order to be better than Life 2. This makes the "better life" relation nonlinear, but we already said that we wish this relation to be quasiorder, therefore it need not necessarily be linear.

We may well have two criteria wherein one criterion has priority over the other. For example, if we write a computer program for a self-driving vehicle, the objective will be less delays and less casualties. We may approach this by expressing delays and human life in monetary value and seeking the least-cost policy. That is, reduce the two criteria to one. This is exactly what we do when we drive a car. We estimate the value of the delay and rush, thus heightening the risk level. So, when we run behind schedule we are prone to take a higher risk of killing somebody. While human drivers can behave like this, your program cannot. As soon as you put a price tag on human life, you will be condemned of inhumanness. That is why you should prioritise the second criterion and compare the two policies on the basis of casualties. When two policies yield the same number of casualties, the one which yields less delays wins. Therefore, we will evaluate human life as infinitely more expensive than delays. This will give us a linear order, but with two criteria.

In this train of thought, when we apply *n* criteria, the score will be an *n*-tuple. The value of each coordinate will be either a real number or the *Undef* constant. The better life will be determined by calculating the arithmetic mean of each coordinate, and the so obtained vector will describe the ultimate result of the whole life. Whether the life is better or not will be determined by comparing the vectors so obtained.

In order to simplify the discussion, in this article we will assume that the criterion is only one. We will note though that Sutton in [2] shoots wide of target by tendering his *"reward hypothesis: all goals and purposes can be well thought of as the maximization of the expected value of the cumulative sum of a single externally received number (reward)"*.

## The discount factor

As mentioned already, the use of a discount factor in the MDP and RL definitions is a mistake. This mistake occurs often when we confuse "what" and "how". What do we want to do and how shall we do it? Well, where would we like to go for a vacation? Our first thought is Honolulu. Then we reckon that Honolulu is too far and on second thought we want a nearer place. This is a mistake, because we messed "what" and "how". What do we want and how shall we achieve it. We want Honolulu and the bare fact that we go to a nearer place does not change the place to which we want to go.

The concept of a discount factor derives from a perfectly natural strategy which tells us that the rewards nearby are more important than those far away. Popular wisdom puts this strategy in a nutshell by the saying "A bird in the hand is worth two in the bush".

The discount factor relates to the strategy rather than to the meaning of life. As from which moment shall we apply the discount factor? That should be the current moment, but it is a moving target, not a fixed time point. If we applied the discount factor as from the initial moment, that would make the beginning of life much more important than its end, which makes little sense.

What was Napoleon's strategy shortly before the battle at Waterloo – "First things first, let's win Waterloo and bother about chores later." So on the eve of the battle it was Napoleon's do-or-die, but if we look at his life as a whole Waterloo does not bear that much weight.

This means that using a discount factor to determine the best of two or more lives is a mistake, at least because the time from which the factor is to be applied and the value of the factor are unknown. If we are less certain we would choose a lower discount factor and hunt for the nearest rewards. The more confident we become, the farther we look in the future and the more prone we are to let the bird in the hand go in order to chase those in the bush. In other words, the more confident we are the more proximal to 1.00 our discount factor will be.

The good thing about the discount factor is that we can use it to evaluate infinite life. This will give us a number which is the sum of a geometric progression. How can we compare two infinite lives and say which one is better? By the following expression:

$$Life1 \geq Life2 \Leftrightarrow \exists n \, \forall (k \geq n) \, begin(Life1, k) \geq begin(Life2, k)$$

Here $begin(Life, k)$ is the beginning of $Life$ the length of which is $k$. The $[\geq]$ symbol is our quasiorder relation which tells us which life is better. We assume that this relation is defined for finite lives and continues it for infinite lives as well. The above formula can also be used to compare a finite and an infinite life, assuming that if $k$ is greater than the life's length then $begin(Life, k) = Life$.

## What is the model as such?

As we said the model of the world will be an oriented graph. Several questions arise. First, is the model of the world deterministic or nondeterministic? Second, what type will be the randomness if we go for nondeterministic graphs? In [5] we discussed two types of randomness. With the first type of randomness we know the exact probability at which each arrow can be chosen (MDP). In the second type of randomness we know the possible arrows, but do not know the probability of choosing each arrow. Figure 5 depicts the second case because the probabilities of same-colour arrows are not indicated. In [5] we also examined a combination of the two randomness types wherein we do not know the exact probability but have an interval and know that the probability is in that interval.

In [5] we demonstrated that with RL the deterministic model and the models with different types of randomness produce equivalent definitions. Therefore, we cannot know whether the world is deterministic or nondeterministic. Of course this implies that we live only one life in the world. If we lived two lives in the world and did the same things each time, the two lives would be identical in the deterministic world and almost surely different in the nondeterministic world.

An important assumption is that with RL we live a single life and have a single history to draw conclusion from. Assuming that we live several times in the same world and gain experience from several histories will produce a substantially different problem where the deterministic and the nondeterministic models are not equivalent.

In this article we will start with a deterministic model, which we will name "perfect model". Then we will define a nondeterministic model or "model with randomness". Then we will replace actions with events and thus obtain an event-driven model.

Where is the score? In this article the score will be part of the observation. In most articles this is not the case. Typically we have two different observations. The first one is named *observation* and indicates the state we are in (Full Observability). The second one is named *score*. The first observation is a label of the state while the second observation is a label of the arrow. In this article we will have a single observation and the score will be part of that observation.

## Incorrect moves

Should we allow for incorrect (bad) moves in the model or assume that each move is correct? For each model with incorrect moves we can construct an equivalent model in which all moves are correct. In [6] we did something similar by building a total model. That is, we added an observation *bad*, which is returned whenever an incorrect move is made.

Let us have two programs for our agent, a Total AI and a Partial AI. The first program will require all moves to be correct, while the second one will permit incorrect moves. If we let the Total AI deal with the world presented as a total model, we will make the agent's task very demanding. In this case the agent will search for patterns in the series of the tried incorrect moves. The agent will not find any pattern, because it does not exist, but will never stop his hunt for patterns. Also, he will try to play an incorrect move twice. Finally the agent will realise that nothing happens by these experiments and will stop making these attempts. This agent will not know that trying an incorrect move does not change anything (other than realising the move's incorrectness, but if he knew that already he does not change anything). These things can also be learnt, but whenever this agent is in a gridlock and ponders what to do he will again and again try things which are not worth trying. Life of the Partial AI in the partial model will be much easier because the Partial AI will be born with some knowledge about incorrect moves and will not sweat to discover and learn this knowledge on its own.

Accordingly, we had better assume that there are incorrect moves. This makes the world greatly simpler and comprehensible. The agent is spared the search of patterns which do not exist and the search for which would be a mere waste of time. Moreover, incorrect moves are a nice example of semi-visible events (these will be discussed in the next sections). Incorrect moves are also a nice example of test states (we discussed them in [5] and will expand in the next article).

## Reinforcement Learning

Now let us provide the formal Reinforcement Learning definition which we will use. We will list what is given and what is to be found. Given are the following elements:

$A$ – the set of the agent's possible actions;
$V$ – the set of possible observations;
$Reward : V \rightarrow \mathbb{R} \cup \{Undef\}$ (a function which for each observation returns a score or *Undef*);
$H$ – history or approximate history of what happened until moment $t$.

Given is also that a perfect model of the world exists and that model is in some current state $s_t$.

What is to be found? We try to answer three questions:
1. What is the world? (establish the model of the world);
2. Where am I now? (determine the current state of the world);
3. What should I do next? (decide our next action as well as our further actions, aiming to maximize the arithmetic mean of scores).

**Note:** This article will deal mainly with questions 1 and 2, while scores are only relevant to the answer of the third question. That is why we will not discuss scores in the next sections.

## The perfect model

Let us now present the perfect model of the world:
$S$ is the set of the internal states of the world,
$s_t$ is the current state of the world, and
$G = <S, R>$ is a total and deterministic oriented graph.
$R \subseteq S \times A \times S$
*View:* $S \to V$
*Incorrect:* $S \to P(A)$

Each edge (arrow) has a label which indicates the action we must make in order to traverse that arrow. Arrows and their labels are determined by the relation $R$. Each vertex (state) has two labels which indicate (i) what we see and (ii) the moves that are incorrect in that state. The labels therefore are (i) the observation in that state and (ii) the set of impossible actions. These labels of the state are determined by the functions *View* and *Incorrect*.

We will assume that the oriented graph $G$ is total and deterministic. Thus, from each state per each action there is an outbound arrow and that arrow is unique. In this case the arrows that represent incorrect moves are somewhat redundant, because we will never use them, but we suppose that they exist as well, because further down we will discuss another model of the world wherein the set of incorrect moves is variable rather than constant. This means that from the same state we may have different incorrect moves at different moments.

Note that if we have a perfect model of the world and know the current state of the world, then the future is completely determinate. So, if we knew the next actions of the agent we would be able to tell exactly what will happen. Unlike the future, the past is not completely determinate because there may be multiple initial states which lead to that current state through history $H$. And even the full history is not determinate, because different initial states may have different full histories. Of course the various full histories must be consistent with history $H$ (they must satisfy the condition *bad* $\subseteq$ *full*). We will assume that there is at least one possible initial state which leads to the current state through history $H$, because we suppose that history $H$ has occurred in that model anyway.

If, in addition to the current state we know the initial state as well, then from the perfect model we can obtain the so-called *abridged model*. In this model, all states and arrows that we have not been traversed through history $H$ will be discarded. In the abridged model we can collect statistics by counting how many times we have traversed each arrow. These statistics will enable us predict the past, the future and the agent's behaviour. By knowing how many times the agent has chosen action A or B in state *s*, we will be able to predict what the agent would do if he finds himself in that state. Indeed, we will try to predict our own behaviour because as we said we will regard the world and the agent as a composite system.

The abridged model may not be perfect. We may come across an arrow which we have never traversed (i.e. an arrow which is absent from the abridged model). Nevertheless, the abridged model presents all we may know because if we never traversed an arrow we have no way to know where it leads to.

## Random variables

Our effort to establish a perfect model of the world is a rather ambitious task which would be feasible only if the world is extremely simple. We are certainly interested in more complex worlds, therefore we can reasonably suggest that the complexity of the world is and will be far beyond full understanding.

Let us assume that there is a limit to our knowledge and that there are things which we will never be able to predict. We will describe them with the term *random variables*. We will regard them as dependencies the complexity of which is beyond our predictive capabilities. For example, if I threw a dice I would expect that it will fall on any number from 1 to 6 with a probability of 1/6. If I had a perfect model I would able to tell the exact number. There is such a perfect model and once I throw the dice I would know which number falls up, but the point is that I need to know it yet before I threw the dice. I cannot know which side will fall up before I threw the dice.

Let me note that if I were extremely smart I would be able to tell which side will be up. If I cannot tell it means that either I am not smart enough or do not have enough information on the basis of which to foretell which number will fall up.

I assume that I am not smart enough and my best guess is a number from 1 to 6 with a probability of 1/6. By saying "my best guess" I do not mean that I will not pursue an even better guess. I might suggest that the dice is skewed and yields 6 with a probability greater than 1/6 or if I blow a breath onto the dice the probability of getting 6 will rise. The last supposition is regarded as superstition (i.e. false), but here we will try to find a statistical dependence between events and if the statistical data indicates that breathing onto the dice works, this will be taken as proof even though it turns out that it has happened by chance.

What is a random variable? Let us suppose I am about to tell you either *black* or *white*. What do you expect to hear – *black* or *white*? I bet you have no idea what exactly I will say. Hence, you expect to hear *white* with a probability in the interval [0, 1].

Imagine I have a dice with one white side and five black sides. You know I will throw that dice and this is how I will decide whether to say *black* or *white*. So you expect *white* with a probability of 1/6. Now imagine I have another dice with two white sides and four black sides. I will throw one of my dices to decide whether I say *black* or *white*. Now you expect to hear *white* with a probability in the interval [1/6, 1/3]. Suppose you know I am more likely to throw the first dice than the second dice. Now you expect *white* with a probability in the interval [1/6, 1/4].

Let me now switch to alternating mode and say *white*, then *black*, then *white*, and so on. Now you expect to hear the opposite of what I said last time. This however is a memory-dependent pattern. I would like to describe non-parameterized random variables. A parameterized random variable is a function which at certain parameters returns a non-parameterized random variable. Later we will describe an oracle α, who depends on the past, on the future and on the current state. Therefore the past is one of the possible parameters.

Suppose when I am in good mood I say *white* or at least I am more likely to say *white*. This adds another parameter to the model, i.e. my mood. So you may try to predict my mood or if it is fully unpredictable then you have a random variable which does not use my mood as a parameter. The variable will nevertheless depend on my mood, but will not be a parameter. It will be a hidden parameter or a parameter which we will not try to predict.

You might suppose that what I am going to tell you is the worst of all worsts. This is a frequently used supposition. In chess playing computer programs, the Min-Max algorithm assumes that the opponent will play the move which frustrates us most. We therefore imagine an opponent whose aim is to hurt us. We do not always anticipate the worst. Sometimes we look forward to the best. E.g. when you come back home from a long journey, what do you expect for dinner? You expect the best because there is someone who loves you and has cooked your most favourite meal.

Let us assume you expect the best. If for some reason *white* is better than *black*, you would expect to hear from me *white*. It may happen that you do not know what is better and will find that out only after the event. Then your expectation will depend on the future. Therefore it will depend on yet another parameter.

## Non-parameterized random variable

So what is a non-parameterized random variable (NPRV)? Most authors suggest that a NPRV has a strictly determined probability (or distribution if there are more than two possible values). Equally this is the supposition in MDP. The MDP chooses randomly the next state, but this randomness is not purely random due to the assumption that the probability of a certain state being selected is precisely defined.

Here we will assume that there are two levels of uncertainty. At the first level we do not know what will happen, but know the exact probability $p$ for it to happen. At the second level of uncertainty we even do not know the probability $p$. It may be that $p$ does not exist at all or it may exist, but we have no way to know it.

When we try to predict a certain event let a certain probability $p$ for this event to occur. If we are smart enough and have the necessary data, we can predict that probability. Fine, but the event will also have a certain value when it occurs and if we are that smart we may even foretell that particular value. At the first level of uncertainty we admit that we cannot foretell the result, while at the second level of uncertainty we go further and admit that we are unable to even foretell the probability.

Can it be that an event has not any exact probability whatever? If it has some probability $p$ and if we observe the event for infinitely long time, the statistically obtained probability should tend to *p* (the Law of large numbers). If we observe the event endlessly, the statistics may not tend to a certain value (we will have limit inferior and limit superior, but they may not coincide).

We can easily imagine a situation where we do not know the exact probability, and all we know is that it is somewhere in the range of *[0, 1]* or *[a, b]*. It is highly difficult to design a physical experiment, which (i) gives us such probability and (ii) that probability is our best guess. Let us for example take a sequence of zeros and ones where limit inferior and limit superior of the arithmetic means are 0 and respectively 1. This sequence will probably move in a zig-zag pattern and when the probability for the last 100 is high we would also expect that it is highly probable for the next number to be 1. Therefore, the probability of that sequence will be in the range *[0, 1]*, but this is not the best description of the random variable, because the last 100 results will give us a better description.

If we wanted to design a physical experiment which produces a precise probability *p*, we can do that easily. All we need is to construct a dice which yields a probability of *p*. We will have the random variable which describes the experiment and, on top of it, it will be the best possible description of the experiment.

How can we then design a physical experiment (i) which yields a probability in the interval *[0, 1]* and (ii) there is no better description of that experiment? Let us assume that the agent plays against a creature which says 1 or 0, but not randomly. The creature seeks to induce confusion in the agent to an extent where the agent is unable to tell whether the next number will be 0 or 1 and, to the worse, is unable to figure out the probability of the next number being 1. We will assume that the creature outsmarts the agent by a wide margin and is very successful in confusing the agent.

The above construct is not an accurate mathematical definition because no matter how smart you are there is always someone smarter. Thus, the creature depends on the agent. If the creature is able to read the agent's thoughts and figure out that the agent expects 1 with a probability in the interval *[a, b]* $\subset$ *[0, 1]*, then the creature can elicit confusion by playing so as to create an appearance that the probability is outside that interval. Reading somebody else's thoughts may sound outlandish, but if the agent is an algorithm (computer program), then the creature can execute that program and find out what the agent's expectations are.

Thus, we have modeled an event the probability of which is in the interval *[0, 1]*. If we aim to design an event the probability of which is in the interval *[a, b]*, then we will construct two dices with probabilities *a* and *b*, and will give these dices to the creature. The creature will chose the dice to throw and still its aim will be to put the agent in confusion.

While the other authors suggest that a non-parameterized random variable has a precise probability (distribution), we will hold that the NPRV has a precise interval of probabilities (or distribution of intervals). Certainly, these probability intervals must satisfy the inequalities described in [5].

We would like to assume that the NPRV does not depend on the past nor of the future. Such NPRV is the dice game, the next number does not depend on the previous throws nor on the future throws. When this assumption comes into play, the order in which the experiments are made will not matter. Consequently, the sequence of results will be immune to permutations. That would be possible only in case we have a fixed probability. If the probability is not fixed, we cannot assume that dependency on the past and on the future does not exist. Accordingly, we will assume that NPRV may depend on the past and on the future, but the dependency is so complex that we are unable to comprehend it. Undetectable dependency is tantamount to non-existent dependency.

## Model with randomness

We suggested that there is unpredictable randomness. We will now construct a second model of the world, which incorporates unpredictable randomness. First, the oriented graph will no longer be deterministic (in the MDP definition the graph is not deterministic, either). Second, the result of our observation in state (*s*) will not be a constant function, but a random variable (observations in POMDP are also random variables). Third, the set of incorrect moves in state *(s)* will also be a random variable. These three random elements will be determined by three oracles ($\alpha$, $\beta$ and $\chi$). The oracles are parameterized random variables and when their parameters acquire a concrete value they become non-parameterized random variables (NPRV).

For each given state $s_t$ and action $a_{t+1}$ oracle $\alpha$ tells which the next state will be. His choices are to a large extent predetermined by graph *G*. Oracle $\alpha$ must choose one of the arrows available. In fact the oracle will have a choice only when the transition is nondeterministic.

For a given state $s_{t+1}$ oracle $\beta$ tells what will be the observation in the moment *t+1*. Here we wrote *t+1* instead of t, because all the three oracles use the same *Past* ($a_{t+1}$ is the end of that *Past*). Oracle $\alpha$ speaks first and says which the next state will be, then oracles $\beta$ and $\chi$ join to say what will be observed in that state ($s_{t+1}$) and which will be the incorrect moves.

Oracle $\chi$ tells whether event *e* occurred at moment *t*. We will use here $\chi$ for events like "a certain move is incorrect". Later we will use $\chi$ to determine other events as well.

Here is the definition of the model with randomness:

*S* is the set of internal states of the world,
$s_t$ is the current state of the world, and
*G* = <*S, R*> is a total oriented graph (nondeterministic).
$R \subseteq S \times A \times S$
$\alpha(Past, s_t, a_{t+1}, Future) \to s_{t+1}$
$\beta(Past, s_{t+1}) \to v_{t+1}$
$\chi(Past, s_{t+1}, e) \to \{true, false\}$

The idea behind our oracles is that the agent does not know which state the world is now in, what will be observed in that state, which the correct moves will be and which will be the next state. The agent does not, but the world knows everything. That is, the world has access to these oracles and is able to tell what exactly will happen. When we have a perfect model of the world (with an initial state), then we know how to define the oracles.

Let us take an arbitrary model *M* (with an initial state). We want to define the oracles for that arbitrary model. The question is whether the *M* model is a model of the world. Any model is a model of the world. Even a model which is not adequate (i.e. does not tell us anything) is still some model of the world.

**Note:** The definition will be based on a single life only. A different life may produce different oracles, but we have only one life.

From *Past* we will derive the step at which we are now (i.e. all we need to know is the length of *Past*). From here we can obtain the values of oracles $\beta$ and $\chi$ for that step. Okay, but to which current state $s_{t+1}$ do these values apply? We may say that they apply to any current state and this definition will be correct, but we prefer them to apply to a single current state (the state which gives us oracle $\alpha$). So, once we define oracle $\alpha$, we can obtain from it oracles $\beta$ and $\chi$.

When model *M* involves nondeterminism, we can define oracle $\alpha$ in various ways. Any such definition will produce some model of the world (although such model may not be adequate). Therefore, any model is determined by (i) an arbitrary oriented graph and (ii) an arbitrary oracle $\alpha$. When we select a different oracle $\alpha$, we will obtain a different trace and in practice a different model. Instead of obtaining the trace from oracle $\alpha$, we can go the other way round. We can select the trace and look for an oracle $\alpha$ which aims to produce this trace. Let us only note that an oracle which produces that trace may not exist, so the oracle obtained will produce a trace that approximates the original trace as closely as possible.

**Note:** Here for each event *e* oracle $\chi$ returns a separate random variable. The definition gives the impression that these are independent random variables, but this may not always be the case. Two events may be related. For example, event *a* may occur only if event *b* has occurred. Another example is where *a* and *b* never occur at the same time. The observation can also be somehow related with the events. It might be appropriate to complicate the definition and merge oracles $\beta$ and $\chi$ in a more complex oracle, but we will not do so.

The oracles depend on the past. We will assume that *Past* is the full history. (We propose to use full history rather than plain history, because oracles depend on what has happened and not on what the agent has observed. The agent knows only the plain history, while the oracle knows the full history. We expect that the agent will learn how to figure out which moves are correct and in practice will also come to know the full history.)

We will assume that oracle $\alpha$ depends on the future as well. This may appear outlandish, but if we consider what is the cause and what is the sequel, the notion that the oracle sees the future is not that bizarre. Let us return to the postman and tramp story. It would be far-fetched to say that the oracle looked into the future, saw some proof of the postman's perpetration appearing in the future and then decided the murder to be committed by the postman. It would make more sense to say the murder was perpetrated by the postman and that is why evidence of the postman's evildoing has appeared in the future.

The oracle's dependence on the future will have some influence on the trace in the model. In Figure 2 we saw that the oracle's choice of arrows leads to a situation where at the next step there is no outbound *b*-event arrow from state 1. So the event b never occurred while we were in state 1. If the oracle's behaviour was different, that would have been a *b*-event arrow from state 1.

We will suppose that the entire future (until the end of life or to infinity) is available to oracle $\alpha$. Nevertheless, what we have in hand is not the entire future, but only the future until the current moment $t$. (The condition of the problem gives the history until that moment $t$.) If we need to know what oracle $\alpha$ will say at moment $t-k$, we have to split history in past (0 to $t-k$) and future ($t-k$ to $t$). We must add that the oracle's decision usually depends only on the near future. In the murder example, if we fail to prove who the murderer is shortly after the murder, probably we will never resolve the murder at all. Once something is proven, there is no way to prove the opposite afterwards.

If we rely on the concept that a life can be lived only once, we can define oracle $\beta$ as a function. Certainly, this function will be defined only for the histories which have played out in that life. As regards the other histories, we will have to think up the values of that function (define them howsoever). If we know the full history of that life, we would be also able to define oracle $\chi$. If we have only the plain history, we will be able to define oracle $\chi$ partially. Oracle $\alpha$ can be defined in any way. We can take a random graph G and define oracle $\alpha$ by that graph. That would define oracle $\alpha$ in all cases with the exception of nondeterministic branches (where we will be at liberty to define the oracle as we wish). An example of a random graph is when we have only one internal state. Even that a graph is one possible model of the world (Figure 1 is such a model).

Note that if we have the same oriented graph, but different oracle $\alpha$, the model will probably be different, because the trace in the model may be different. We will look for different traces in the model rather than for different oracles $\alpha$. If two oracles $\alpha$ produce identical traces in the model, we will deem that these oracles are equivalent. Conversely, if we have the trace, we can reverse-engineer from it an oracle who induces that trace (if such an oracle ever exists, of course). Here is our reverse engineering process: from all possible choices of the oracle we will remove those which are not possible for the particular trace and for the given future. Then, if there are any choices left, we will select one of them (with the probability corresponding to the probability of this choice being possible with the given trace and future).

## The trace

Our idea is that the oriented graph *G* describes the world somehow. Here it turns out that a random graph *G* can be a model of the world! The truth is that any graph can be a model, but very few graphs are adequate models of the world. When we are in some of the graph states, we expect something to happen. We expect that certain events will never occur in that state or vice versa – certain events must occur. An event may occur with a probability much higher or much lower than the expected probability for that event. If nothing happens in any state, then the model will be completely inadequate.

In our terminology, the occurrences during our journey along the arrows form the *trace* in the model. The perfect model gives us a very clear trace. In the perfect model, at each state there is a precisely defined observation and a precisely defined set of incorrect moves. In the abridged version of the perfect model some arrows will not be there. A missing arrow will mean that a certain event never occurred in a certain state (a certain move was never played). This is also a trace. The question is how to interpret it? Should we assume that this event will never happen in this state, or that it may happen albeit with a very limited probability.

An event may not necessarily depend only on your immediate observations or actions. It may also depend on the more distant history, for example on your observation at the previous step.

From where can we glean traces? From statistics. We will take the abridged model (that is, we will remove all arrows and states that have not been used until the present moment). We will count how many times each arrow has been used. For each state and each event we will count how many times the event occurred in that state. Of course we cannot count that for each and every event, because the events are infinitely many. We will only take count of certain more important events. From these statistics we can identify excursions from the expected value and these excursions will be the traces we are looking for.

Note that the statistical exercise is doable by the world, but a very challenging task for the agent. The world knows in which state it is and can count the arrows, while the agent can only make guesses. Even with in perfect model the agent may not know the state in which he has been (may not know the current state, but even if he knew that knowledge this does not determine the pervious states uniquely).

Sometimes the agent does not know his current state and can find that out only afterwards. That would not be a problem, because statistics will be compiled with time lag. But, it will be a problem if the agent never figures out the exact state in which he has been. Then the gathering of statistics becomes a very demanding task.

## The exhaustive model

Let us assume that we have a model in which oracles $\alpha$, $\beta$ and $\chi$ do not depend on the past. This will be referred to as an *exhaustive model*, i.e. a model which cannot be improved any further. Everything that needs to be remembered about the past is already memorized in the current state. The model can be improved if we split one state in two and in a certain history we come to one of these states with a different probability. That is, the two states are distinguishable from the viewpoint of the past. In the exhaustive model, the two states will not be distinguishable from the viewpoint of the future, because the oracles are independent from the past. This renders such splitting of states meaningless.

There may be an exhaustive model with only one state. The world in this case is terrible and past occurrences are of no consequence for the future. It does not matter at all which action we choose in that world. Certainly, such a world is overly simplistic. We suggest that the worlds we are interested in are much more complicated and it is virtually impossible to establish perfect or exhaustive models of these worlds.

We will not attempt to achieve a perfect understanding of the world (establish a perfect model) or an understanding to the level of unsolvable randomness (establish an exhaustive model). We will try to establish a variety of multiple models which describe various features of the world and objects in the world. The Cartesian product of all these models will give us a model which probably will not be exhaustive either, but will describe the world reasonably well.

The comprehensiveness of the world is known as *Markov property*. In RL it is typically assumed that there is an exhaustive model of the world. Here we suggested something more. We suggested that there is a perfect model of the world.

However, here we will search for adequate models rather than for exhaustive models. Thus, we will look for models that have some trace. Trying to establish an exhaustive model is a superfluously ambitious problem and even an absolutely unsolvable problem in the case of more complex worlds.

## Visible events

An event will be a Boolean function which at any point of time is true or false. First we will say what is a visible event: an event, which can be seen from the history (withal not from the whole history, but from its end).

**Definition:** A visible event will be the set of local histories. This event will be true when the history ends in some local history from the set of local histories.

An example of a visible event is our last move. An example of an event which is not visible is whether a move is correct. The correctness of the move can be seen from the full history, but cannot be seen from the plain history. This type of event will be termed *semi-visible*, because it contains in it (as a subset) one visible event. The visible event contained in it is: "This move is incorrect and we have already tried it".

Another example of a semi-visible event is sunrise. You may well see sunrise but can also miss it if you overslept. The fact that you overslept sunrise does not mean there was no sunrise at all. There surely was, but you simply did not see it.

An example of an invisible event is: "I caught a cold". You cannot eyewitness the occurrence of that event, but various signs afterwards will tell you that it has occurred (high body temperature, etc.).

So far the set of events which we used to find the model of the world was the set of our actions. Now we will generalize and use a random set of events.

## The Event-Driven model

Now we will replace actions with events. We will take a set of events, and assume that the set is not big because if the events are far too many the model will become excessively complicated.

In the new model the arrows will no longer be labeled with actions, but with events. In the case of actions, two actions could not take place at the same time. When we have events however, it is perfectly possible that two events occur at the same time. That is why oracle $\alpha$ will not be defined by an action, but by events, and indeed not by a single event, but by the *events* (which is a set of events). When *events* is the empty set, then we do not go anywhere else and $\alpha$ returns the same state $s_t$. If in *events* there is only one event, the oracle will choose one of the possible arrows of that event. Where the *events* set contains two events, the oracle will have to decide whether to choose one of them (as if one event has obscured the other) or assume that the two events occurred one after the other. That is, traverse an arrow labeled with the first event and then another an arrow labeled with the second event. Moreover, the oracle should decide which event occurs first and which one occurs second.

$S$ is the set of internal states of the world,
$s_t$ is the current state of the world,
$E$ is a set of events, and
$G = <S, R>$ is a total oriented graph (nondeterministic).
$R \subseteq S \times E \times S$
$\alpha(Past, s_t, events, Future) \to s_{t+1}$
$\beta(Past, s_{t+1}) \to v_{t+1}$
$\chi(Past, s_{t+1}, e) \to \{true, false\}$

Besides the incorrect moves, here oracle $\chi$ determines all invisible events in E, and the invisible part of the semi-visible events. As regards visible events, they depend only on *Past* so the workings of the oracle are sufficiently clear. That is why it is superfluous for the oracle to recognize visible events. We do not want the oracle to recognize visible events, because we have an exhaustive model when oracles do not depend on *Past*. To avoid changing the exhaustive model definition, we will assume that $\chi$ recognizes only invisible events, while the visible events are known.

The event-driven model also lets us extract an abridged version with the arrows that have been used and count how times they have been used. Again, only the world can obtain an exact count, while the agent can only make a rough estimation of that count.

From those statistics we can derive the trace in the model and thereby judge the adequacy of the model.

## The model with variables

It is perfectly reasonable to include variables in the model. In a certain state, one event may keep occurring for some time and then stop occurring. A variable is a convenient presentation of whether the event is occurring or not. That will be a local variable because it is pertinent to a certain state. Nothing prevents us from having global variables which are pertinent to multiple states.

An example of a model with variables was provided in [5]. There we had many doors (states) and each door was either locked or unlocked. Therefore, there was one variable attached to each door.

$S$ is the set of internal states of the world,
$Var$ is the set of variables,
$s_t$ is the current state of the world,
$eval_t$ is the current evaluation of the variables, and
$G = <S, R>$ is a total oriented graph (nondeterministic).
$R \subseteq S \times E \times S$
$\alpha(Past, s_t, eval_t, events, Future) \to <s_{t+1}, eval_{t+1}>$
$\beta(Past, s_{t+1}, eval_{t+1}) \to v_{t+1}$

$\chi(Past, s_{t+1}, eval_{t+1}, e) \rightarrow \{true, false\}$

From a theoretical perspective, we have not altered anything here, because by including variables we only changed the number of states. This can also be seen above. Wherever we had a state before, we have a state and an evaluation of the variables now.

Although in theory the inclusion of variables does not change anything, in practice things change a lot, because if we have 10 Boolean variables, the number of states expands by a factor of 1024, which is a major increase. If we try to chart an oriented graph with that many states we will end up with something immensely complicated. In the end of the day, variables describe the trace (what happens in that state). The values of most variables will be completely unknown to us.

## The Cartesian model

As mentioned already, we are looking for various models which describe various features of the world. Our model of the world, i.e. the model we have discovered, will consist of all these models. We can reckon that it is a Cartesian product of all these models.

In which state shall we be? It will be the current states of all these models and the current evaluations of the variables in these models. We can perceive the current state as a tuple of all these current states and current evaluations of variables.

You are now in your home city. The day is Monday. The time is 10 am. You are full because you had a steady breakfast. Each of these four sentences describes an event-driven model and the current state of that model.

At this moment you see the computer screen. This is an object (an event-driven model) and you can see it, so you are not in the *outside* state.

At this moment the door of your room is unlocked and the front door of the building is locked. So you have in your mind a model of the building and can tell whether two of the doors are locked or unlocked. Therefore, you know the values of two variables in this model.

We will assume that none of the models discovered is an exhaustive one, because if there is one exhaustive model, all other models are redundant. While a Cartesian product of multiple non-exhaustive models may return an exhaustive model, this is very unlikely if the world is sufficiently complicated.

## Agents and objects

We should make a distinction between agents and objects because these are two different entities.

In previous articles [7] we referred to agents. He who changes something is an agent and we must predict his future actions to find out if he is friend or foe and try to come to terms with him.

What is man, an agent or an object? On the one hand, a man an agent, because he changes the world. A man can move things around or even steal things. On the other hand, a man is an object, because we recognize (identify) him when we see him, hear him speaking on the phone or hear his name pronounced by somebody else.

## Conclusion

In this article we discussed the relation between events and their sequels. (The sequel is what we call a *trace* in the event-driven model.) The question is, when does an event occur? For example, when does sunrise occur – when the sun peeks a little bit, when it is halfway up or when it is all up? Here we selected the moment in which we deem that the event has occurred and this is the moment when we understood that the event has occurred. It is by no means obligatory for the sequels of that event to commence exactly from that moment. When the sun rises, and even before it shows up, it sheds light all over and right away. The sun also sheds warmth, but much later, not right away. When we search for a trace we should be mindful that there may be sequels of the event or sequels of the fact that we come to a certain state of the model, and should not expect the sequel to emerge outright. We must allow for the sequel to play out slightly before or after. Moreover, the sequel (trace) will not be related to the states of the model only. We may as well see some sequels of the event proper. At sunrise for example the sky becomes red at about the time when the sun shows up. This is not related to the state before or after sunrise (i.e. to the night or to the day).

Therefore, when we gather statistics we should consider how close we are to the event and search for a trace (traits) both in the periods between events and around the events as such. The trace is a trait which occurs in a certain state of the model, but can also occur when we traverse some arrows in the model.

The trace will not consist exclusively of events. Besides events, we will have tests, and objects. Tests are special types of event, which we will highlight below. The appearance of an object is also an event and that event can help us define the trace in an event-driven model. That is, for defining an object we can use other objects. For example, in one room we see a cat and this is the room with the cat. This is how the cat object helps us define and remember a whole room. Of course the cat is an object which possesses the property *mobility*. We may have to search for the cat throughout the house. If the cat had the property *uniqueness*, once we find it in a room we will no longer search other rooms. The cat may not necessarily stay in one room, but in different rooms we stand different probabilities of seeing the cat. A different probability can also be a trace in the model (in this case the model is the house).

We saw so far that whatever events and whatever oriented graph we grab, it is always a model of the world. That bad news is that if the model is picked randomly, it will almost certainly turn out to be inadequate. In other words, it will have no trace whatever and nothing special (nothing interesting) will occur in its states.

In some cases (as in Figure 2) the model will have a trace, but that trace will only come from oracle $\alpha$ and we will be unable to use it (in this case to predict the next event) because we cannot know the state in which are.

It emerges therefore that the problem of discovering an adequate model of the world is very challenging. For this purpose we must search for specificities. We should keep observing various events and be alert enough to note when a specific combination of events occurs. For example, a black hat and untidy hair – that must be our colleague John.

Very few are the events which we can observe on a permanent basis. We will observe most events only from time to time. We call these events *tests* [5]. Tests are crucial for discovering adequate event-driven models.

The next article will be dedicated exactly to tests and how they can help us efficiently discover event-driven models.

## References

[1] Richard Sutton, Andrew Barto (1998). Reinforcement Learning: An Introduction. *MIT Press, Cambridge, MA (1998)*.

[2] Richard Sutton (2008). Fourteen Declarative Principles of Experience-Oriented Intelligence. www.incompleteideas.net/RLAIcourse2009/principles2.pdf

[3] Johan Åström. (1965). Optimal control of Markov processes with incomplete state information. *Journal of Mathematical Analysis and Applications. 10: 174–205*.

[4] Apostolos Burnetas, Michael Katehakis. (1997). Optimal Adaptive Policies for Markov Decision Processes. *Mathematics of Operations Research. 22 (1): 222*.

[5] Dimiter Dobrev (2017). How does the AI understand what's going on. *International Journal "Information Theories and Applications", Vol. 24, Number 4, 2017, pp.345-369*.

[6] Dimiter Dobrev (2018). Minimal and Maximal models in Reinforcement Learning. *August 2018. viXra:1808.0589*.

[7] Dimiter Dobrev (2008). The Definition of AI in Terms of Multi Agent Systems. *December 2008. arXiv:1210.0887*.